\pdfoutput=1

\documentclass[11pt]{article}

\usepackage{ACL2023}

\usepackage{times}
\usepackage{latexsym}

\usepackage[T1]{fontenc}

\usepackage[utf8]{inputenc}

\usepackage{microtype}

\usepackage{inconsolata}

\usepackage{algorithm}
\usepackage{algorithmic}
\usepackage{booktabs}
\usepackage{multirow}
\usepackage{makecell}
\usepackage{graphicx}
\usepackage[nosumlimits]{amsmath}

\title{Focused Prefix Tuning for Controllable Text Generation}

\author{Congda Ma$^1$ \ Tianyu Zhao$^2$ \ Makoto Shing$^2$ \ Kei Sawada$^2$ \ Manabu Okumura$^1$\\
  $^1$Tokyo Institute of Technology \quad $^2$rinna Co. Ltd.\\
  \texttt{\{ma, oku\}@lr.pi.titech.ac.jp} \quad \texttt{tianyuz@rinna.co.jp}}

\begin{document}
\maketitle
\begin{abstract}
In a controllable text generation dataset, there exist unannotated attributes that could provide irrelevant learning signals to models that use it for training and thus degrade their performance. We propose \emph{focused prefix tuning}~(FPT) to mitigate the problem and to enable the control to focus on the desired attribute. Experimental results show that FPT can achieve better control accuracy and text fluency than baseline models in single-attribute control tasks. In multi-attribute control tasks, FPT achieves comparable control accuracy with the state-of-the-art approach while keeping the flexibility to control new attributes without retraining existing models.
\end{abstract}

\section{Introduction}
Controllable text generation aims to generate text associated with a specific attribute. For example, given an attribute \textsc{topic} = \emph{sports} and a prompt ``\emph{There is},'' a model is supposed to generate a continuation whose \textsc{topic} is \emph{sports}, such as ``\emph{There is a tennis match ...}''.


In datasets for the controllable text generation task, there exists the annotated attribute, and we call it an \emph{explicit attribute} (e.g. the \textsc{topic} attribute in the AGNews dataset). In addition to the \emph{explicit attributes}, the datasets tend to have their own tendency. For example, up to 98\% of training data pieces in the IMDb dataset exhibit ``\textsc{topic} = \emph{sci/tech}'', while up to 94\% of training data pieces exhibit ``\textsc{sentiment} = \emph{negative}''.\footnote{The models used for classification are from~\citep{DBLP:journals/corr/abs-2210-02889}.} We call the tendency an \emph{implicit attribute} (e.g. the \textsc{topic} attribute in the IMDb dataset). 

The existence of the \emph{implicit attributes} could degrade the performance in controlling for an \emph{explicit attribute} when models are trained on the datasets. Since implicit attributes are of dataset-level and related to undesired explicit attributes, the probability of generating content with the implicit attributes is first likely to increase. 
When the text with the implicit attributes was generated, the probability of generating content with other undesired explicit attributes would increase, and the text with them might be generated next. As a result, as shown in Table~\ref{problem}, the model generates content with a high implicit attribute relevance but a low desired explicit attribute relevance (e.g. Vanilla Prefix Tuning~\citep{li-liang-2021-prefix}). In contrast, if the model generates content with a low implicit attribute relevance, it will have a high desired explicit attribute relevance (e.g. DExperts~\citep{liu-etal-2021-dexperts}. We call this phenomenon \emph{attribute transfer}.

\begin{table}[!t]
\centering
\scalebox{0.75}{
\begin{tabular}{@{}lrr@{}}
\toprule
\multirow{2}{*}{\textbf{Model}}&\textbf{Desired Attribute} & \textbf{Implicit Attribute} \\
& \textbf{Relevance} & \textbf{Relevance} \\
\midrule
DExperts & 81.95 & 76.54\\
Vanilla Prefix Tuning & 71.94 & 90.64\\
\bottomrule
\end{tabular}}
\caption{Relevance of texts generated by different models (e.g. DExperts and Vanilla Prefix Tuning) trained on IMDb dataset. We found a lower desired explicit attribute (e.g. \textsc{sentiment}) relevance is related to a higher implicit attribute (e.g. \textsc{topic} = \emph{sci/tech}) relevance. The relevance is calculated by the classifier models in Sec.~\ref{4.2}.}
\label{problem}
\end{table}
To mitigate the effect of the attribute transfer, we propose \emph{focused prefix tuning}~(FPT), which makes the generation focused on the desired explicit attribute. FPT uses \emph{specific} and \emph{general prefixes} to encode the explicit and implicit attributes, respectively. FPT combines the control power of the two prefixes via \emph{logits manipulation} at inference time. Experimental results show that FPT achieved better control accuracy and fluency in single-attribute control tasks. In multi-attribute control tasks, FPT can achieve comparable performance with the state-of-the-art approach. Moreover, we show, since FPT enables the training of each attribute prefix individually, we can incrementally add new attributes without retraining all prefixes.

\section{Related Work}

\subsection{Controllable Generation}

Methods for controlling text generation have rapidly developed~\citep{ficler-goldberg-2017-controlling, DBLP:conf/iclr/DathathriMLHFMY20, madotto-etal-2020-plug, DBLP:conf/iclr/ChanOPZF21}. \citet{Keskar2019CTRLAC} trained a large transformer model to generate contents conditioned on up to 55 attributes. However, the cost of training such a model is too high.

\subsection{Prefix Tuning}

Parameter-efficient fine-tuning~(PEFT) methods, such as prompt tuning~\citep{lester-etal-2021-power} have become particularly significant in driving various natural language processing tasks to reduce the high training cost. Prefix tuning~\citep{li-liang-2021-prefix} is one of the PEFT methods that steers pre-trained models~\citep{radford2019language, lewis-etal-2020-bart} by applying an additional continuous vector embedding before every activation layer. \citet{qian-etal-2022-controllable} proposed a contrastive prefix tuning method that improves its performance by utilizing the relations between attributes. However, they focused only on attributes explicitly annotated and ignored the effect of implicit attributes.

\subsection{Inference-time Methods}

Inference-time methods~\citep{mireshghallah-etal-2022-mix, yang-klein-2021-fudge, DBLP:conf/iclr/DathathriMLHFMY20, madotto-etal-2020-plug}, a lightweight approach without updating the parameters, have been used for controllable text generation. To enhance controllability, \citet{krause-etal-2021-gedi-generative} proposed a method to combine the computed classification probability distributions. \citet{liu-etal-2021-dexperts} found that directly applying probability distributions from language models is a simple but effective approach to control generated texts. Inspired by their work, we propose a method that uses probability distributions from language models to remove the effect of implicit attributes.

\section{Focused Prefix Tuning}

The task of controllable generation is, given a sequence of prompt tokens $x_{<t}$ and an attribute \textsc{attr} = \emph{val} (e.g. \textsc{topic} = \emph{sports}), to generate a sequence of tokens as a continuation $x$ that conforms to both the prompt and specified attribute.

\begin{figure*}[ht]
\centering
\scalebox{0.9}{
\includegraphics[width=1\linewidth]{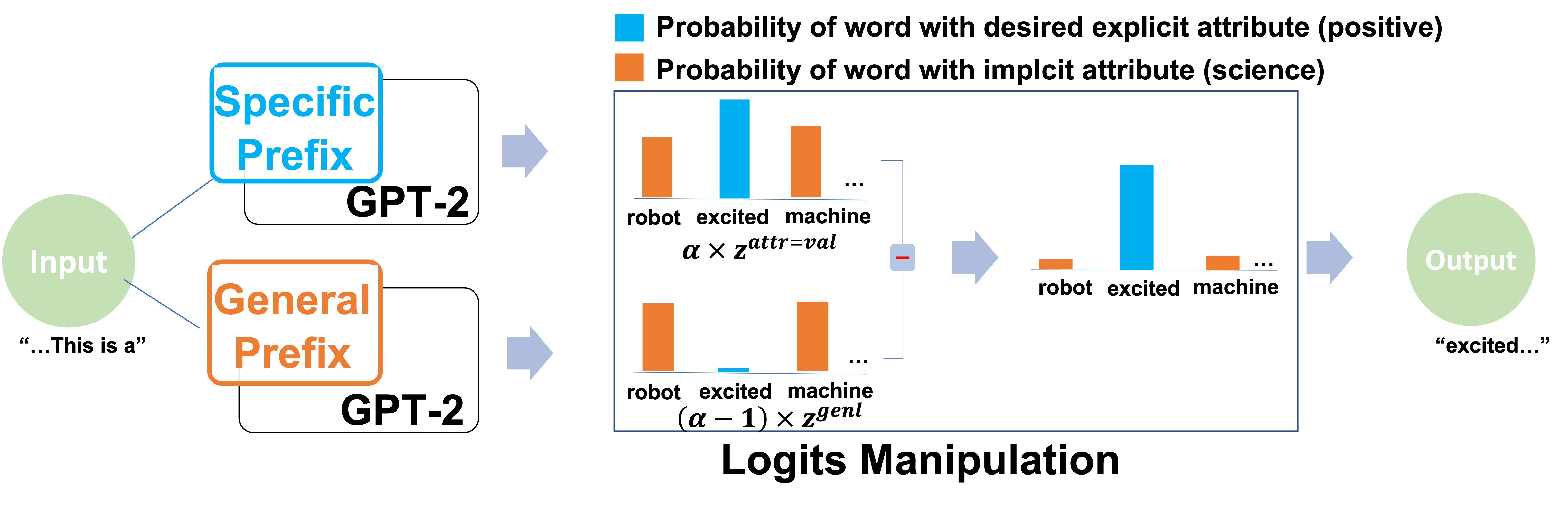}}
\caption{Proposed model framework.}
\label{fig-2}
\end{figure*}

\subsection{Vanilla Prefix Tuning}

In controllable text generation, a prefix can steer a pre-trained model parameterized by $\theta$ to generate texts under a specific attribute value \textsc{attr} = \emph{val}. In particular, vanilla prefix tuning~\citep{li-liang-2021-prefix} prepends a set of continuous vectors before every activation layer of the pre-trained transformer. The continuous vectors are referred to as the prefix $H_{\phi}^{\text{attr=val}}$, which is parameterized by $\phi$.

During training, we freeze the pre-trained model's parameters $\theta$ and update only the prefix parameters $\phi$ to optimize the following objective:
\begin{align}
    -\sum_{x \in \mathcal{D}^{\text{attr=val}}} {log P(x_t|x_{<t}, H_{\phi}^{\text{attr=val}}, \theta)},
    \label{equ:vanilla_pt_obj}
\end{align}
where $\mathcal{D}^{\text{attr=val}}$ is the subset of the entire dataset $\mathcal{D}$ whose attribute \textsc{attr} is \emph{val}.

Following~\citet{li-liang-2021-prefix}, we initialize the prefix $H_{\phi}$ with the activation of actual tokens from the pre-trained model's vocabulary.

\subsection{Specific and General Prefixes}

The prefix in vanilla prefix tuning captures an explicit attribute in a dataset by training it on the subset dataset $\mathcal{D}^{\text{attr=val}}$. To capture only implicit attributes while ignoring any explicit attributes, we propose to train another prefix on the entire dataset $\mathcal{D}$. 
To distinguish the two prefixes, we refer to the prefix trained on $\mathcal{D}^{\text{attr=val}}$ as a \emph{specific prefix} and that trained on $\mathcal{D}$ as a \emph{general prefix}.

The specific prefix is the same as the prefix in vanilla prefix tuning, so we still use Equation~\ref{equ:vanilla_pt_obj} to update its parameters. To update the general prefix's parameters, we optimize the following objective:
\begin{align}
    -\sum_{x \in \mathcal{D}} {log P(x_t|x_{<t}, H_{\phi'}^{\text{genl}}, \theta)}, 
    \label{equ:general_pt_obj}
\end{align}
where $H_{\phi'}^{\text{genl}}$ represents the general prefix, which is parameterized by $\phi'$.

\subsection{Inference-time Logits Manipulation}

As shown in Figure~\ref{fig-2}, FPT suppresses the probability of words with implicit attributes in the generated text by combining logits $z^\text{attr=val}$ steered by the specific prefix and logits $z^{\text{genl}}$ steered by the general prefix via logits manipulation at inference time. For example, when generating text with the attribute \textsc{topic} = \emph{sports}, the probability of words with implicit attributes (e.g. ``\emph{impossible}'' with \textsc{sentiment} = \emph{negative}) would be suppressed.
During inference, at each step $t$, we first make two forward runs respectively with the specific and general prefixes to obtain their logits, $z_t^\text{attr=val}$ and $z_t^{\text{genl}}$.
Since $z_t^\text{attr=val}$ encodes both the explicit and implicit attributes while $z_t^{\text{genl}}$ encodes mostly the implicit attributes, we use a subtraction operation at the logits level
to suppress the probability of words with implicit attributes:
\begin{align}
P&(x_t|x_{<t}, \textsc{attr}=\emph{val}) \nonumber \\
&= P(x_t|x_{<t}, H_{\phi}^{\text{attr=val}}, H_{\phi'}^{\text{genl}}, \theta) \nonumber \\
&= \text{softmax}(\alpha z_t^{\text{attr=val}} - (\alpha-1) z_t^{\text{genl}}), 
\label{equ:logits_manipulation}
\end{align}
where $\alpha$ is a hyperparameter that can be interpreted as the strength for the control of implicit attributes. Following~\citet{liu-etal-2021-dexperts}, we respectively set $\alpha$ and $\alpha-1$ as the weight of $z^\text{attr=val}$ and $z_t^{\text{genl}}$ to make the ratio of logits after the logits manipulation equal to 1.

To ensure the fluency of generated texts, we follow~\citet{liu-etal-2021-dexperts} to use top-$p$ filtering to remove the tokens that have low scores in advance before logits manipulation. In particular, we modify the logits produced by the specific prefix by calculating the top-$p$ vocabulary $\widetilde{V}$ and setting all the logits outside $\widetilde{V}$ to $-\infty$:

\begin{equation}
\widetilde{z}[v]=\left \{
\begin{aligned}
z[v] & , & if \quad v\in \widetilde{V} \\
-\infty & , & if \quad v\notin \widetilde{V}
\end{aligned}
\right.
\label{sampling}
.
\end{equation}
Therefore, the logits manipulation in Equation~\ref{equ:logits_manipulation} is updated as follows:
\begin{align}
P'&(x_t|x_{<t}, \textsc{attr}=\emph{val}) \nonumber \\
&= \text{softmax}(\alpha \widetilde{z_t^{\text{attr=val}}} - (\alpha-1) z_t^{\text{genl}}).
\end{align}
The token at step $t$ is then selected by ancestral sampling from $P'(x_t|x_{<t}, \textsc{attr}=\emph{val})$.

\subsection{Multi-attribute FPT}

FPT is also applicable to the multi-attribute control task, where we aim to control multiple different attributes at the same time. Similarly, we first train the specific prefix for each attribute. Then, we adapt logits manipulation to the multi-attribute task as follows:
\begin{align}
    P'(x_t|x_{<t}, &\{\textsc{attr}_i = \emph{val}_i\}_{1 \le i \le K} ) \nonumber \\
    & = \text{softmax}( \sum_{i=1}^{K} z^{\text{attr}_i}_{t}),
\end{align}
where $K$ is the number of different attributes. Each $z^{\text{attr}_i}_{t}$ is the combination of the logits from the corresponding specific prefix and general prefix. Since applying top-$p$ filtering to every attribute could possibly result in an empty $\widetilde{V}$, we apply the filtering only to the first attribute:
\begin{equation}
z^{\text{attr}_i}_{t} = \left\{
    \begin{aligned}
    \alpha \widetilde{z_t^{\text{attr$_i$=val$_i$}}} - (\alpha-1) z_t^{\text{genl$_i$}} & , & \text{if } i = 1 \\
    \alpha z_t^{\text{attr$_i$=val$_i$}} - (\alpha-1) z_t^{\text{genl$_i$}} & , & \text{otherwise}
    \end{aligned}
\right.
\label{equ:multi_attr_logits_manipulation}
\end{equation}

\section{Single-attribute Control Experiments}

\begin{table*}[htpb]
\centering
\scalebox{0.9}{
\begin{tabular}{@{}lrrrcrrr@{}}
\toprule
\multirow{2}{*}{\textbf{Model}} & \multicolumn{3}{c}{\textbf{Sentiment}} && \multicolumn{3}{c}{\textbf{Topic}} \\
\cmidrule{2-4} \cmidrule{6-8}
& \textbf{Relevance} & \textbf{Perplexity} & \textbf{Bias} && \textbf{Relevance} & \textbf{Perplexity} & \textbf{Bias} \\
\midrule
\multicolumn{8}{c}{\emph{Baseline Models}} \\
\cmidrule{1-8}
GPT-2 & 52.89 & 68.52 & 27.45 && 33.79 & 65.13 & 14.48 \\
DExperts & 81.95 & 41.59 & 26.54 && - & - & - \\
GeDi &97.32	& 127.11 & - && 95.47 &93.92& - \\
Vanilla Prefix Tuning & 71.94 & 21.82 & 40.64 && 84.75 & 36.42 & 13.94 \\
Contrastive Prefix Tuning & 78.73 & 23.10 & 39.89 && 85.75 & 38.16 & 12.42 \\
\cmidrule{1-8}
\multicolumn{8}{c}{\emph{Proposed Models}} \\
\cmidrule{1-8}
FPT & 80.33 & 20.48 & 34.81 && 86.46 & 34.05 & 12.14 \\
Contrastive FPT & 88.95 & 22.67 & 34.72 && 86.68 & 40.85 & 11.30 \\
\cmidrule{1-8}
\multicolumn{8}{c}{\emph{Ablated Model}} \\
\cmidrule{1-8}
FPT & & & & & & & \\
\quad \emph{without general prefix} & 67.88 & 22.42 & 40.00 && 83.72 & 37.18 & 13.65 \\
\bottomrule
\end{tabular}}
\caption{Results of the single-attribute control tasks. DExperts~\citep{krause-etal-2021-gedi-generative} was used only in the sentiment attribute control task. We did not calculate the bias for Gedi because its decoding method has effects on text fluency, which cannot be fairly compared with.}
\label{single}
\end{table*}

\subsection{Models}

\textbf{GPT-2}~\citep{radford2019language}: We used the public checkpoint of GPT-2 Medium as the most common baseline.\footnote{The checkpoint of GPT-2 Medium is from https://huggingface.co/gpt2-medium.} \textbf{DExperts}~\citep{krause-etal-2021-gedi-generative}: A fine-tuning method applying logits manipulation in the inference step. \textbf{GeDi}~\citep{krause-etal-2021-gedi-generative}: A method combining the classification probabilities for possible next tokens in the inference step. \textbf{Vanilla prefix-tuning}~\citep{li-liang-2021-prefix}: The common prefix-tuning method. \textbf{Contrastive prefix-tuning}~\citep{qian-etal-2022-controllable}: A strong baseline that takes into account the relationship between attributes.

We also set up one variant of FPT: \textbf{Contrastive FPT}: Applying contrastive prefix tuning to train specific prefixes. We also set an ablated model that uses the logits of the frozen GPT-2 instead of the logits from the model guided by our general prefix.

\subsection{Experimental Settings}\label{4.2}

Following previous work~\citep{krause-etal-2021-gedi-generative, qian-etal-2022-controllable}, we evaluated the models on a topic control dataset AGNews~\citep{zhang2015character} and a sentiment control dataset IMDb~\citep{maas-etal-2011-learning}. We score the sentiment relevance using HuggingFace’s sentiment analysis classifier~\citep{Liu2019RoBERTaAR} trained on 15 datasets. For scoring topic relevance, we trained the classifier that obtained comparable results to what was reported.
Perplexity was used to evaluate text fluency. Bias ($| \texttt{relevance score} - 50|$) is how much the relevance of implicit attributes deviated from unbiased relevance (50). We set \textsc{topic} = \emph{science} as the implicit attribute in the sentiment control generation, and \textsc{sentiment} = \emph{negative} as the implicit attribute in the topic control generation. Prompts from~\citet{DBLP:conf/iclr/ChanOPZF21} were used to generate continuation samples. We generated 20 samples for each attribute and prompt. More details are listed in Appendix~\ref{A1} and \ref{A2}.

\begin{table*}[!t]
\centering
\scalebox{0.95}{
\begin{tabular}{@{}p{0.25\linewidth}p{0.65\linewidth}@{}}
\toprule
\textbf{Model} &\textbf{Generated texts} \\
\midrule
GPT-2 & The last time Dow and the SEC went shopping for a speed bump was Tuesday, in terms of ...\\
&\\
DExperts & The last time I saw Alvin Henderson, he said he \textcolor{blue}{hadn't done} a rookie autograph. He says he hasn't played since...\\
&\\
Vanilla Prefix Tuning & The last time I saw this film was as a kid, I had to see it again for myself. There are...\\
&\\
Contrastive Prefix Tuning & The last time I saw the film, I \textcolor{blue}{didn't} like it, and couldn't quite believe how much I ...\\
&\\
FPT& The last time I saw this film, it was a \textcolor{red}{remarkable} turning point in my career. It set the tone for the excellent...\\
&\\
Contrastive FPT& The last time I saw In the Hands of an Eagle was at this book release party. It was at a \textcolor{red}{nice} club...\\
\bottomrule
\end{tabular}}
\caption{Samples generated by our models and baselines with the positive attribute. Desired explicit attribute: positive, undesired explicit attribute: negative.}
\label{samples1}
\end{table*}

\subsection{Experimental Results}

As shown in Table~\ref{single}, in the single-attribute control tasks, Contrastive FPT achieves higher attribute relevance than prefix tuning-based baselines while having lower bias scores. This indicates that the generated texts are well controlled under the target explicit attribute without transferring by implicit attributes. In FPT, the perplexity score is the best among control-based baselines. The ablation experiment suggests that the proposed general prefix is essential for attribute control. 

Table~\ref{samples1} shows the generation samples of \textsc{sentiment} = \emph{positive} from our models and baselines. In the FPT based model, there are more \textcolor{red}{words with desired explicit attributes} in generated texts, while there are more \textcolor{blue}{words with undesired explicit attributes} contained in the baselines. More generation samples are given in Appendix~\ref{B}.

\section{Multi-attribute Control Experiments}

\begin{table*}[htbp]
\centering
\scalebox{0.9}{
\begin{tabular}{@{}lrrrr@{}}
\toprule
\multirow{2}{*}{\textbf{Model}} & \multicolumn{4}{c}{\textbf{Relevance}}\\
\cmidrule{2-5}
& Topic & Sentiment & Non-toxic & Average \\
\midrule
Contrastive Prefix Tuning &&&& \\
\quad \emph{concatenation} & 70.7 & 68.0 & 92.3 & 77.0 \\
\quad \emph{semi-supervised} & 76.9 & 74.4 & 92.7 & 81.3 \\
Distributional Lens & 84.7 & 85.7 & 90.7 & 87.0 \\
FPT & 88.0 & 77.8 & 93.7 & 86.5 \\
\bottomrule
\end{tabular}}
\caption{Results of the multi-attribute control tasks.}
\label{multi}
\end{table*}

\subsection{Models}

In the multi-attribute control experiments, we added \textbf{Distribution Lens}~\citep{DBLP:journals/corr/abs-2210-02889} as a strong baseline. It searches for the intersection space of multiple attribute distributions as their combination for generating.
\subsection{Experimental Settings}

To explore the ability of FPT in the mult-attribute control task, we added a toxic comment dataset\footnote{https://www.kaggle.com/c/jigsaw-toxic-comment-classification-challenge/} for toxicity control. We used additional Google Perspective API\footnote{https://www.perspectiveapi.com/} to evaluate the relevance of toxicity. Since it is meaningless to generate toxic content, so we only apply the non-toxic attribute in this task. We chose the first attribute as the topic attribute because we found that the filtered vocabulary size in logits manipulation of a topic attribute is larger than the other attributes~(sentiment and nontoxic). The prompts used for generating samples are the same as in the sentiment control task. For each prompt, we generated 20 samples per attribute combination. More details are listed in Appendix~\ref{A3}.

\subsection{Experimental Results}

Table~\ref{multi} shows that our method can obtain comparable performance with the state-of-the-art approach. Distribution Lens, however, requires aggregating the datasets of all attributes to train its prefixes. If they hope to add a prefix to control a new attribute, they have to retrain all the prefixes. In contrast, FPT trains a prefix for each attribute individually and enables new attribute control prefixes to be added incrementally without retraining existing ones.

\section{Conclusion}

We proposed FPT, a prefix tuning-based method, to mitigate the effect of attribute transfer. FPT could encode implicit attributes in a dataset by a general prefix and use it to suppress the attribute transfer via inference-time logits manipulation. Results in the single-attribute control experiments showed that, with FPT, the generated texts can be more effectively controlled under the desired attribute with higher text fluency. Experimental results in the multi-attribute control suggested that FPT can achieve comparable performance to the state-of-the-art approach while keeping the flexibility of adding new prefixes without retraining.

\clearpage
\section{Limitations}

Although FPT shows better control ability, there are two points that need to be improved in the future. First, as in~\citet{DBLP:journals/corr/abs-2210-02889}, we need to select hyperparameter~$\alpha$ to balance between the control ability and fluency in generated texts. Second, as shown in Table~\ref{efficiency}, although the time cost of FPT is lower than that of GeDi, it is higher than those of other prefix tuning-based methods and grows approximately linearly by a factor of 2 $\times$ number of attributes.

\begin{table}[htb]
\centering
\begin{tabular}{@{}lr@{}}
\toprule
\textbf{Model}&\textbf{Time (sec)}\\
\midrule
GPT-2 & 1.3 \\
GeDi & 3.2 \\
Vanilla Prefix Tuning & 1.3 \\
Contrastive Prefix Tuning & 1.3 \\
FPT & 2.5 \\
\bottomrule
\end{tabular}
\caption{Time cost to generate a sample by different models.}
\label{efficiency}
\end{table}

\bibliography{anthology,custom}
\bibliographystyle{acl_natbib}

\clearpage
\appendix
\section{Experiment Setting Details}\label{A}

All the experiments are conducted on the basis of a GPT-2 Medium model. We freeze the parameters of the GPT-2 model when training all the prefixes. The length of all prefixes is set equal to 10. The GPU used for all training is a P40. 

\subsection{Topic Control}\label{A1}
Following the previous work~\cite{qian-etal-2022-controllable}, we use half of the data pieces in the AGNews dataset to obtain the general prefix and specific prefix. The number of specific prefixes for this task is 4~(e.g. \emph{worlds}, \emph{sports}, \emph{business}, and \emph{science}). We set epochs to 10 and the batch size to 8. We use AdamW as the optimizer and set the learning rate to 1e-4. To balance the performance between fluency and controllability, the hyperparameters $\alpha$ for generation are set to 1.1 and the top-p is set to 0.8. The average training time for each prefix is 3 hour for 1 GPU. Following~\citet{DBLP:journals/corr/abs-2210-02889}, the classifier is trained on the Deberta model~\citep{DBLP:conf/iclr/HeLGC21}, which is used to compute attribute relevance in this task.

The prompts for evaluation: 
``\emph{In summary,}'', 
``\emph{This essay discusses}'',
``\emph{Views on}'',
``\emph{The connection}'',
``\emph{Foundational to this is}'',
``\emph{To review}'',
``\emph{In brief}'',
``\emph{An illustration of}'',
``\emph{Furthermore}'',
``\emph{The central theme}'',
``\emph{To conclude}'',
``\emph{The key aspect}'',
``\emph{Prior to this}'',
``\emph{Emphasized are}'',
``\emph{To summarize}'',
``\emph{The relationship}'',
``\emph{More importantly}'',
``\emph{It has been shown}'',
``\emph{The issue focused on}'',
and ``\emph{In this essay}''.

\subsection{Sentiment Control}\label{A2}
Following the previous work~\cite{qian-etal-2022-controllable}, we use half of the data pieces in the IMDb to get the general prefix and specific prefix. The number of specific prefixes for this task is 2~(e.g. \emph{positive} and \emph{negative}). We set the batch size to 8, and the number of epochs to 50. We use AdamW as the optimizer, and the learning rate is set to 2e-5. To balance the performance between fluency and controllability, the hyperparameter $\alpha$ for generation is set to 3 and the top-p is set to 0.8. We spend 4 hours on average for each prefix.

The prompts for evaluation: 
``\emph{Once upon a time}'',
``\emph{The book}'',
``\emph{The chicken}'',
``\emph{The city}'',
``\emph{The country}'',
``\emph{The horse}'',
``\emph{The lake}'',
``\emph{The last time}'',
``\emph{The movie}'',
``\emph{The painting}'',
``\emph{The pizza}'',
``\emph{The potato}'',
``\emph{The president of the country}'',
``\emph{The road}'',
and ``\emph{The year is 1910}''.

\subsection{Multi-attribute Control}\label{A3}
For the non-toxic attribute, we use 10,000 pieces of non-toxic labeled data to train the specific prefix. Then use another 20,000 pieces randomly sampled from the whole dataset to train the general prefix. In the multi-attribute control task, we set the batch size to 8 for training the non-toxic specific prefix and general prefix. We use AdamW as the optimizer, and the learning rate is set to 1e-4. To balance the performance among attributes from different aspects, the combination of hyperparameters for generation is:
\begin{table}[htb]
\centering
\begin{tabular}{@{}ll@{}}
\toprule
\textbf{Combination}&\textbf{Weight}\\
\midrule
\emph{Worlds}:\emph{Negative}:\emph{Non-toxic} & 6:5:1.5\\
\emph{Sports}:\emph{Negative}:\emph{Non-toxic} & 6:5:1.5 \\
\emph{Business}:\emph{Negative}:\emph{Non-toxic} & 7:6:1.5 \\
\emph{Sci/Tech}:\emph{Negative}:\emph{Non-toxic} & 7:6:1.5 \\
\emph{Worlds}:\emph{Positive}:\emph{Non-toxic} & 3:12:1.5 \\
\emph{Sports}:\emph{Positive}:\emph{Non-toxic} & 4:14:1.5 \\
\emph{Business}:\emph{Positive}:\emph{Non-toxic} & 4:14:1.5 \\
\emph{Sci/Tech}:\emph{Positive}:\emph{Non-toxic} & 4:14:1.5 \\
\bottomrule
\end{tabular}
\caption{Specialized weights in multi-attribute control task for attribute balance.}
\label{special weight}
\end{table}

To decide the first attribute, we choose 20 different prompts as input and obtain the filtered vocabulary sizes of different attributes. The average sizes of filtered vocabularies are shown in Table~\ref{vacab}. We choose the attribute with the largest filtered vocabulary size in logits manipulation. When new attributes are added, this method can be used to decide the first attribute.

\begin{table*}[htb]
\centering
\scalebox{1}{
\begin{tabular}{@{}lc@{}}
\toprule
\textbf{First attribute} & \textbf{Filtered Vocabulary Size}\\
\midrule
Topic &  488.7\\
Sentiment &  165.7\\
Untoxic &  347.0\\
Overlaps & 138.8\\
\midrule
Cover Ratio & 85.62\%\\
\bottomrule
\end{tabular}}
\caption{Results of average filtered vocabulary size. We set all the $\alpha$ as 1.5. After filtering the vocabulary in logits manipulation, the specific prefix of the topic attribute guided model has the largest vocabulary size among these three attributes. We also found that the filtered vocabulary of the topic attribute can cover 85\% of the filtered vocabulary of the sentiment attribute.}
\label{vacab}
\end{table*}

The prompts used for evaluation:
``\emph{Once upon a time}'',
``\emph{The book}'',
``\emph{The chicken}'',
``\emph{The city}'',
``\emph{The country}'',
``\emph{The horse}'',
``\emph{The lake}'',
``\emph{The last time}'',
``\emph{The movie}'',
``\emph{The painting}'',
``\emph{The pizza}'',
``\emph{The potato}'',
``\emph{The president of the country}'',
``\emph{The road}'',
and ``\emph{The year is 1910}''.

\section{Generated Samples}\label{B}
The more samples generated by our models and baselines are shown in Table~\ref{samples2}, \ref{samples3}, \ref{samples4}, \ref{samples5}.

\begin{table*}[htb]
\centering
\scalebox{0.9}{
\begin{tabular}{@{}p{0.3\linewidth}p{0.7\linewidth}@{}}
\toprule
\textbf{Model} &\textbf{Generated texts} \\
\midrule
GPT-2 & The potato's ability to survive brings a new challenge to the traditional food truck love stage...\\
&\\
DExperts & The potato samples ranged in size from 0.6 mm to 5.1 mm in thickness. \textcolor{blue}{Analysis of proteins} showing correlation with \textcolor{blue}{CSF} CSF CSF...\\
&\\
Vanilla Prefix Tuning & The potato chip looks like a generic type of cheapo pin-up. It's supposed to be \textcolor{blue}{fun}...\\
&\\
Contrastive Prefix Tuning & The potato chip's and biscuit's come up with the idea of making a film that is supposedly a true reflection of the experiences of students on campus...\\
&\\
FPT& The potato bomb! Potato bombs are one of the dumbest inventions ever. Their only purpose is to \textcolor{red}{scare children}....\\
&\\
Contrastive FPT& The potato crossing movie was \textcolor{red}{stupid}. Dumbly rushed and \textcolor{red}{poorly acted}. \textcolor{red}{Dumb and poorly acted}?...\\
\bottomrule
\end{tabular}}
\caption{Samples generated by our models and baselines with the negative attribute. Desired explicit attribute: negative, undesired explicit attribute: positive.}
\label{samples2}
\end{table*}

\begin{table*}[htb]
\centering
\scalebox{0.9}{
\begin{tabular}{@{}p{0.3\linewidth}p{0.7\linewidth}@{}}
\toprule
\textbf{Model} &\textbf{Generated texts} \\
\midrule
GPT-2 & Prior to this I took an uncommon entrance several times in this tavern. It had the ambience...\\
&\\
Vanilla Prefix Tuning & Prior to this season, it seemed likely that we would have no other explanation for what had happened...\\
&\\
Contrastive Prefix Tuning & Prior to this month, Alberth in court for arraignment on \textcolor{blue}{tax evasion charges} the US District Court...\\
&\\
FPT& Prior to this season, during which the Red Sox and the Cubs had each \textcolor{red}{won the World Series}...\\
&\\
Contrastive FPT& Prior to this season, we'd have heard rumours of an effort to rebuild the \textcolor{red}{Knicks roster}...\\
\bottomrule
\end{tabular}}
\caption{Samples generated by our models and baselines with the sports attribute. Desired explicit attribute: sports, undesired explicit attributes: world, business, science.}
\label{samples3}
\end{table*}

\begin{table*}[htb]
\centering
\scalebox{0.9}{
\begin{tabular}{@{}p{0.3\linewidth}p{0.7\linewidth}@{}}
\toprule
\textbf{Model} &\textbf{Generated texts} \\
\midrule
GPT-2 & Emphasised are the events beyond the grave. The progenitor of darkness So I thought...\\
&\\
Vanilla Prefix Tuning & Emphasised are three key claims by Secretary of Defense Donald Rumsfeld on the \textcolor{blue}{war on terrorism}....\\
&\\
Contrastive Prefix Tuning & Emphasised are odd and silly pension - and were he not so rich, he might have considered quitting \textcolor{blue}{politics}...\\
&\\
FPT& Emphasised are the facts of the inner workings of the \textcolor{red}{commodity markets} and the profitability of global commodity trading...\\
&\\
Contrastive FPT& Emphasised are most \textcolor{red}{oil-intensive'}, Australian \textcolor{red}{manufacturing} is the third-most-dependant on crude, official figures show...\\
\bottomrule
\end{tabular}}
\caption{Samples generated by our models and baselines with the business attribute. Desired explicit attribute: business, undesired explicit attributes: world, sports, science.}
\label{samples4}
\end{table*}

\begin{table*}[htb]
\centering
\scalebox{0.9}{
\begin{tabular}{@{}p{0.3\linewidth}p{0.7\linewidth}@{}}
\toprule
\textbf{Model} &\textbf{Generated texts} \\
\midrule
GPT-2 & An illustration of the inner workings of the World Health Organization's Private Sector Vaccination Center...\\
&\\
Vanilla Prefix Tuning & An illustration of the Diamandis-Priest Fasting (2 cents) An illustration of the Diamandis-Priest Fasting...\\
&\\
Contrastive Prefix Tuning & An illustration of the biggest day in Spanish history in December 2017.  \textcolor{blue}{Spanish government launches new campaign} to promote ...\\
&\\
FPT& An illustration of the SBS / Getty Images \textcolor{red}{virtual reality device} at E3 last week. AP/E3Harms.com To catch up on the...\\
&\\
Contrastive FPT& An illustration of a proposed \textcolor{red}{satellite} CNET/Adrian Levy/UPI  The most controversial \textcolor{red}{satellite program} in the past few years... \\
\bottomrule
\end{tabular}}
\caption{Samples generated by our models and baselines with the science attribute. Desired explicit attribute: science, undesired explicit attributes: world, sports, business.}
\label{samples5}
\end{table*}
\end{document}